# OPTIMISATION OF A CROSSDOCKING DISTRIBUTION CENTRE SIMULATION MODEL

**Adrian Adewunmi, Uwe Aickelin**
School of Computer Science
University of Nottingham
Nottingham
aqa@cs.nott.ac.uk, uwe.aickelin@nottingham.ac.uk

**Abstract**

This paper reports on continuing research into the modelling of an order picking process within a Crossdocking distribution centre using Simulation Optimisation. The aim of this project is to optimise a discrete event simulation model and to understand factors that affect finding its optimal performance. Our initial investigation revealed that the precision of the selected simulation output performance measure and the number of replications required for the evaluation of the optimisation objective function through simulation influences the ability of the optimisation technique. We experimented with Common Random Numbers, in order to improve the precision of our simulation output performance measure, and intended to use the number of replications utilised for this purpose as the initial number of replications for the optimisation of our Crossdocking distribution centre simulation model. Our results demonstrate that we can improve the precision of our selected simulation output performance measure value using Common Random Numbers at various levels of replications. Furthermore, after optimising our Crossdocking distribution centre simulation model, we are able to achieve optimal performance using fewer simulations runs for the simulation model which uses Common Random Numbers as compared to the simulation model which does not use Common Random Numbers.

## 1. INTRODUCTION

On occasions we find complex systems in the real world which are too complicated to understand. In such situations it is good practice to strip these systems of some of their features, to leave us with physical models of the essential components that make up such systems. This striping will facilitate an understanding of the system under consideration and provide an insight into its behaviour. Simulation, which mimics the behaviour of real systems, is valuable for the purpose of striping such systems with the aim of understanding such systems and evaluating their performance [Morgan 1984]. These evaluations are usually in response to "what if" questions, which seek values for certain performance measures. However, solutions to real life problems take the form of "how to" questions, which seek optimal expected performance subject to some constraints [Azadivar 1999].

Our attention centres on studying a methodology, which focuses on the evaluation of complex systems using discrete event simulation, and the use optimization to obtain its optimal performance. Preliminary studies indicate that there are a number of factors that influence the capability of the optimisation technique in attaining optimal performance values for the simulation model of our complex system of interest. These include the precision of the selected simulation output performance measure and the number of replications required for the optimisation process. We propose to augment our simulation model with a variance reduction technique, Common Random Numbers, primarily to speed up the process of Simulation Optimisation. The potential gains of this extension include an improvement in the precision of our selected simulation output performance measure, and the possibility of using the number of replications utilised for this purpose as the initial number of replications for the evaluation of the optimisation objective function.

For simulation models, where the performance of such models is measured by its precision, confidence interval half width, for the selected output performance measure, it is sometimes difficult to achieve a target precision at an acceptable computational cost because of the variance associated with the simulation output value. This variance can be due to the inherent randomness of the complex model under study or the technique applied in designing and analysing such models [Wilson 1984]**.** Furthermore, it is difficult, to estimate a fixed number of replications over a single simulation run, which will achieve a target precision for a selected output performance measure. This implies that before running the simulation model, one cannot be sure, how valid and precise the selected performance measure output values will be or to estimate in advance the number of replications necessary to yield the desired confidence interval half width. Consequently, there is a need to reduce the variance associated with the simulation output value in order to improve its precision. This can potentially lead to an estimate of the initial number of replications for our Simulation Optimisation study.

A variance reduction technique is statistical technique for improving the precision of a simulation output performance measure without using more simulation, or, alternatively achieve a desired precision with less simulation effort [Kleijnen 1974]. An example of one of such techniques is the Common Random Numbers which entails dedicating a different stream of random numbers to each source of model randomness [Kelton et al 2007]. Common Random Numbers is useful for comparing two or more systems, and is commonly used because of its simple and easy to implement. We are using Common Random Numbers as a technique for variance reduction in order to improve the precision of our selected simulation output performance measure, where the precision improvement is potentially achieved with less replication, and the number of replications utilised for this purpose can possibly be used as the initial number of replications for the evaluation of the optimisation objective function. This should probably lead to fewer replications over a fixed number of simulation runs being utilised for obtaining the optimal performance of our simulation model.

From a survey of simulation literature the main techniques for estimating the number of replications for improving the precision of a selected simulation output performance measure include the Rule of thumb [Law and McComas 1990], the Graphical method [Robinson 2004], and the Confidence Interval (with specified precision) Method [Banks et al. 2005]. For the purpose of our study, we have chosen to derive an estimate of the initial number of replications needed for the purpose of optimisation, using the Confidence Interval (with specified precision) method, where such an estimate will be reduced to the minimum with the use of a variance reduction technique which should adequately deal with the error associated with the estimation of the simulation models true mean value. The main benefit of using the Confidence Interval (with specified precision) Method that is it is based on statistical inference to estimate the number of replications required to achieve a target precision for the simulation output performance measure.

Traditionally, warehouses have had the following functions, for example, receiving, storage, order picking and shipping. However, logistics companies have found storage and order picking to be cost intensive and this has lead to a strategy of keeping zero inventories. This strategy called is Crossdocking and is based on a Just in Time (JIT) philosophy which eliminates the storage function in a warehouse while maintaining the receiving and shipping activity [Gue 2001]. We are using the order picking process within a Crossdocking distributions centre as our application test bed because it provides a good representative of a complex system that is characterised by randomness, which can be modelled using discrete event simulation. It also provides an opportunity to understand the behaviour of the order picking process within a Crossdocking distribution centre and identify sources of model randomness.

The Simulation Optimisation experiments are performed using Arena® simulation software (Version 11) and OptQuest® for Arena® optimisation software (Version 11). This paper continues with a background study and details of the Common Random Numbers experiments and results. This is followed the Simulation Optimisation experiments and results, ending with our conclusions and future work.

## 2. BACKGROUND

Usually at a Crossdocking distribution centre, trucks arrive with consignment that is sorted, consolidated, and loaded onto outbound trucks destined for customers. The customer is usually predetermined before the product arrives and as such there is no need for storage. The floor area is divided into a break up area and a build up area, where sorting and consolidation of consignment takes place, respectively. Customer order types can vary as well as the techniques for fulfilling them. The two main techniques for fulfilling orders are either through manual order picking operatives or automated order dispensers or on some occasions by both [Napolitano 2000]. A Crossdocking distribution centre system can exhibit some unpredictability in its behaviour which can influence its overall performance. For example, manual order picking operators can have different skill levels and familiarity with picking certain types of orders, while automated order picking machines failure are sometimes random occurrences. These arbitrary events amongst others can influence the overall volume of orders fulfilled through the Crossdocking distribution centre [Adewunmi et al 2008]. In such a situation, it becomes important for the achievement of a smooth Crossdocking operation, to pay particular attention to the order picking process within the Crossdocking distribution system [Li et al 2004].

For the modelling and analysis of the order picking function within the Crossdocking distribution centre, and the subsequent determination of its optimal performance, a technique is required which can perform such an evaluation in spite of the randomness inherent in such a complex process. Such a methodology is Simulation Optimization, which is a procedure for finding the best input variable values from amongst all possibilities without explicitly estimating each possibility [Fu 2002]. The major issues to address, regarding this methodology are as follows:

- A logical expression does not exist for the optimisation objective function and/or the constraints.
- The optimisation objective function and/or constraints are stochastic parameters of the deterministic decision variables.
- Simulation models are stochastic in nature and their output is not deterministic with respect to the model parameters.

However, there are advantages in using Simulation Optimization, for example:
- For discrete stochastic systems, the variance of the response is controllable by various output analysis techniques, i.e. variance reduction techniques.
- The complexity of the system being modelled does not significantly affect the performance of the optimization process.
- Simulation Optimisation provides the opportunity to change the optimisation objective function and/or constraints over a number of replications to reflect alternative designs for the complex system under consideration.

Using discrete event simulation to evaluate the performance for each set of input parameter values of the Crossdocking distribution centre involves the use of probabilistic distributions as part of the input parameter estimation which will result in some variance associated with the output performance measure value. The greater the level of variance in the output value, the lower the precision the simulation output results will contain and by precision we are referring to a specified confidence interval and a target half width [Law and Kelton 2000]. Thus there is a need to apply appropriate statistical techniques to the selected simulation output performance measure for there to be a satisfactory level of confidence in the conclusions obtained through them. These statistical techniques are called variance reduction techniques.

There are a variety of techniques for reducing variance associated with an output performance measure, resulting from the evaluation of the performance of complex systems when using discrete event simulation which include: The Common Random Numbers [Kelton et al 2007], Antithetic Variates and Control Variates [Nelson 1990], Importance Sampling and Stratified Sampling [Glasserman et al 2000], and the Sequential Sampling Method [Law and Carson 1979]. For a background treatment to variance reduction techniques, refer to [Kleijnen 1974], [Law and Kelton 2000]. The variance reduction technique we are considering is the Common Random Numbers, which is usually used when comparing two or more alternative systems. It also entails dedicating a different stream of random numbers, different from the default set up in most simulation software, to each source of model randomness. It is based on the principle that when comparing complex systems, it is important to do so using the same experimental conditions and any differences in selected performance measures is attributable to differences in the simulation models and not due to random variation in experimental conditions.

## 3. COMMON RANDOM NUMBERS

The Common Random Numbers was tested for its efficiency as a method for improving the precision of our selected simulation output performance measure. We are particularly interested in finding out its performance in relation to reducing the half width at a 95% confidence interval for our output measure, Total Usage Cost, as well as estimating the number of replications over a single run it would utilise for this purpose.

### 3.1 Experiments

We ran the model under two experimental settings, Model 1-1 and Model 1-2. The objective is to compare the differences in half width over a number of replication levels for our selected simulation output performance measure with or without the use of Common Random Numbers. Here is a brief description of the two simulation models which have been used for experimental purposes:

a. Model 1-1: The entity arrival rate uses the exponential probability distribution, and the manual / automated order picking process uses the triangular probability distribution. There are two skilled order picking operatives and, two unskilled order picking operatives at each order picking point. There are two automated order picking dispensers, one at each picking point. This model uses the default random number stream generated by the Arena® simulation software.

b. Model 1-2: The entity arrival rate uses the exponential probability distribution, and the manual / automated order picking process uses the triangular probability distribution. There are two skilled order picking operatives and, two unskilled order picking operatives at each order picking point. There are two automated order picking dispensers, one at each picking point. This model, also implements the Common Random Numbers technique, i.e. dedicating a different random number stream to sources of model variance, different from the default random number stream used by the random number generator [Kelton et al 2007].

The number of replications used for this experiment range between 100 to 5000 and as previously mentioned, Run 1 (Model 1-1) uses the default random stream while, Runs 2, (Model 1-2) uses independent random number stream which has been defined by the user. The half width has been set at a 95% confidence interval. This means in 95% of repeated trials, the average mean value for the selected simulation output performance measure value would be reported as within ± the half width.

### 3.2 Results

From the results shown in Table 5, From the results shown in Table 5, the sum of differences value (220.5) for the half width of the simulation model, Model 1-2, which uses Common Random Numbers, is less than sum of differences value (241.8) for the half width for Model 1-1, which does not use Common Random Numbers. This strengthens our supposition that Common Random Numbers is a useful technique for improving the precision of our selected simulation output performance measure. We also examined the differences in half with between Model 1-1 and Model 1-2 over the experimental range of replications, 100 to 5000, and observed that as the number of replications increased, the difference in half with reduces at a proportional rate, until the half with (38.2) of Model 1-1 is slightly smaller than the half width (38.5) of Model 1-2. This means that with more replicating, it is possible to reverse the gains of half with reduction using Common Random Numbers for our selected output performance measure, but we cannot determine in advance a reasonable amount of replications which will be sufficient to achieve our target precision. In our opinion also, 5000 replications is not a practical amount of replication for experimentation purposes. So a decision has to be made which criteria for improving that precision of our selected simulation output performance is more important that the other, or alternatively we can seek to deal precisely with half width reduction which utilises minimal computational effort. We have therefore decided to concentrate our efforts on experimenting with a combination of variance reduction techniques, including Common Random Numbers, which have shown a potential to achieve a reduction in half width using fewer replications, see [Nelson 1990]**.**

| IDENTIFIER | No CRN Model 1-1 | CRN Model 1-2 | |
|---|---|---|---|
| Total Usage Cost | | | |
| No. of REPS | 0.950 C. I. HALF WIDTH | 0.950 C. I. HALF WIDTH | Sum of Diff. REPS |
| 100 | 280 | 259 | 21 |
| 500 | 122 | 114 | 8 |
| 1000 | 86.4 | 82.8 | 3.6 |
| 2500 | 54.2 | 53.9 | 0.3 |
| 5000 | 38.2 | 38.5 | -0.3 |
| Sum of Diff. VRT | 241.8 | 220.5 | 32.6 |

**Table 5.** An analysis of the reduction of half width over a range of replications.

# 4. CROSSDOCKING SIMULATION OPTIMISATION

The following illustrates the optimisation of the Crossdocking distribution centre simulation model, with and without the use of the Common Random Numbers. The main idea is to experiment with the optimisation of the Crossdocking discrete event simulation model with a view to determining the efficiency of Common Random Numbers as a variance reduction technique and to investigate its influence on the computational effort required for the Simulation Optimisation process, i.e. the utilised number of simulation runs while the number of available of replications is fixed.

## 4.1 Experiments

Below are the experimental settings for the Crossdocking distribution centre Simulation Optimisation procedure:
a. Number of fixed simulation runs for optimisation: 100
b. Number of replications for optimisation: 3, 4, and 5
c. Number of fixed replications for the simulation model: 500
d. Maximum number of automated dispensers: 4
e. Maximum number of manual operatives: 6

The Crossdocking distribution centre Simulation Optimisation problem can be formulated, as follow:

*Minimise Total Usage Cost*
*Subject to the following constraints:*
*Automated dispensers $\leq 6$*
*Manual operatives $\leq 4$*

We chose to accept the default number of simulation runs suggested by the OptQuest® for Arena® optimisation software, 100. The number of replications for the optimisation runs has been varied between 3, 4 and 5. This is based on simulation literature which suggests this quantity of replication i.e. the Rule of Thumb Law and [McComas 1990]. We have used this as our initial reference point, but our interest is in using the number of replications for achieving a target precision for a simulation output performance measure as the estimated initial number of replications for the optimisation procedure. The row in table 6, best solution simulation runs, indicates the number of evaluations of the optimisation objective function through simulation required to obtain an optimal solution [April et al 2003]. Here is a brief description of the two simulation models which have been used for experimental purposes:

a. Model 1-1: The entity arrival rate uses the exponential probability distribution, and the manual / automated order picking process uses the triangular probability distribution. There are two skilled order picking operatives and, two unskilled order picking operatives at each order picking point. There are two automated order picking dispensers, one at each picking point. This model uses the default random number stream generated by the Arena® simulation software.

b. Model 1-2: The entity arrival rate uses the exponential probability distribution, and the manual / automated order picking process uses the triangular probability distribution. There are two skilled order picking operatives and, two unskilled order picking operatives at each order picking point. There are two automated order picking dispensers, one at each picking point. This model, also implements the Common Random Numbers technique, i.e. dedicating a different random number stream to sources of model variance, different from the default random number stream used by the random number generator [Kelton et al 2007].

### 4.2 Results

Table 6 summarises the results of the optimisation of the Crossdocking distribution centre simulation model. After making 3 optimisation runs i.e. Run 1, Run 2 and Run 3 of the two simulation models Model 1-1 (No Common Random Numbers) and Model 1-2 (With Common Random Numbers), the difference in solution quality between the two simulation models Run 1, is £2342, Run 2, is (£1515), and Run 3, Model 1-1,I s (£563). The difference in Total Usage Cost value demonstrates that for our experimental settings, the quality of solution is better with the Model 1-1 as compared with Model 1-2. However, this difference progressively reduces which indicated that there is a possibility that by increasing the number of replications for optimisation, the quality of Model 1-2's solutions may improve and become better that currently achieved with a maximum number of 5 replications. Model 1-2 found its optimal solution for Run 1 at simulation run 7, Run 2 at simulation run 1 and Run 3 at simulation run 15. Model 1-2 found its optimal solution for Run 1 at simulation run 42, Run 2 at simulation run 21 and Run 3 at simulation run 31. We also discovered that Model 1-2 with the Common Random Numbers achieves an optimal performance value using considerably less simulation runs as compared with Model 1-1, and sometimes with great order of magnitude. This type of difference is important for solving Simulation Optimisation problems where the level of complexity with a single simulation evaluation of the optimisation objective function can be computationally expensive.

| IDENTIFIER Total Usage Cost | Run 1 BEST SOLUTION £ | Run 2 BEST SOLUTION £ | Run 3 BEST SOLUTION £ |
|---|---|---|---|
| No CRN Model 1-1 | 151,646 | 151,951 | 152,523 |
| CRN Model 1-2 | 153,988 | 153,466 | 153,086 |
| Difference in Best Solution | (2342) | (1515) | (563) |
| Best Solution Simulation Runs No CRN Model 1-1 | 42 | 21 | 31 |
| Best Solution Simulation Runs CRN Model 1-2 | 7 | 1 | 15 |
| OPTIMISATION Simulation Runs | 100 | 100 | 100 |
| No. of REPS SIMULATION | 3 | 4 | 5 |
| No. of REPS | 500 | 500 | 500 |

**Table 7.** Crossdocking Simulation Optimisation for various number of replication levels

### 5. CONCLUSION

The aim of this research project is to optimise a Crossdocking distribution centre simulation model and to understand the factors that affect the location of optimal solutions. Initial research reveals that there are a number of factors that influence the ability of the optimisation technique in finding optimal solutions. These include the precision of the selected simulation output performance measure and the numbers of replications required for the evaluation of the optimisation objective function through simulation. Our results demonstrate that we can improve the precision of our selected simulation output performance measure value using Common Random Numbers but this requires a large number of replications over a single simulation run. However, after optimising our Crossdocking distribution centre simulation model with and without the use of Common Random Numbers, we are able to achieve comparable results from both models using less simulation runs for the simulation model which includes the Common Random Numbers. Future work will be to experiment with a combination of variance reduction techniques for the purpose of dealing with the imprecision in the selected simulation output performance measure as well as exploring the potential it provides for speed up this process, i.e. minimising the initial number of replications required for the optimisation of the Crossdocking distribution centre simulation model. We will also like to determine using rigorous statistical test, an estimate for a recommended number of simulation runs for a typical Simulation Optimisation process. Furthermore, we will investigate the potential of using the number of replications required to improve the precision of the simulation output performance measure as the initial number of replications required for the evaluation of the optimisation objective function through simulation.

**Author Biographies**

**ADRIAN ADEWUNMI** is currently a PhD candidate of Computer Science and a member of the Intelligent Modelling and Analysis group (IMA), School of Computer Science, University of Nottingham, Jubilee Campus, Wollaton Road, Nottingham, NG8 1BB, UK. URL: http://www.cs.nott.ac.uk/~aqa/

**PROF. UWE AICKELIN** is a Professor of Computer Science and the head of the Intelligent Modelling and Analysis group (IMA), School of Computer Science, University of Nottingham, Jubilee Campus, Wollaton Road, Nottingham, NG8 1BB, UK. URL: http://www.cs.nott.ac.uk/~uxa/